\documentclass[10pt]{article}
\usepackage[letterpaper]{geometry}
\usepackage{hicss}
\usepackage{times}
\usepackage[none]{hyphenat}
\usepackage{url}
\usepackage{latexsym}
\usepackage{minted}
\usepackage{indentfirst}
\usepackage{graphicx}
\usepackage{array}
\graphicspath{{images/}}
\usepackage[
    style=apa,
  ]{biblatex}
\usepackage{comment}

\title{CognitiveSky: Scalable Sentiment and Narrative Analysis for Decentralized Social Media}

\author{Gaurab Chhetri \\
Department of Computer Science \\
Texas State University \\
San Marcos, Texas, USA\\
{\underline{ gaurab@txstate.edu}} \\ \And
Anandi Dutta, Ph.D. \\
Ingram School of Engineering \\
Texas State University \\
San Marcos, Texas, USA\\
{\underline{ anandi.dutta@txstate.edu}} \\ \And
Subasish Das, Ph.D. \\
Ingram School of Engineering  \\
Texas State University \\
San Marcos, Texas, USA\\
{\underline{ subasish@txstate.edu}} \\
}

\date{06-14-2025}

\begin{document}
\maketitle
\begin{abstract}
The emergence of decentralized social media platforms presents new opportunities and challenges for real-time analysis of public discourse. This study introduces CognitiveSky, an open-source and scalable framework designed for sentiment, emotion, and narrative analysis on Bluesky, a federated Twitter or X.com alternative. By ingesting data through Bluesky’s Application Programming Interface (API), CognitiveSky applies transformer-based models to annotate large-scale user-generated content and produces structured and analyzable outputs. These summaries drive a dynamic dashboard that visualizes evolving patterns in emotion, activity, and conversation topics. Built entirely on free-tier infrastructure, CognitiveSky achieves both low operational cost and high accessibility. While demonstrated here for monitoring mental health discourse, its modular design enables applications across domains such as disinformation detection, crisis response, and civic sentiment analysis. By bridging large language models with decentralized networks, CognitiveSky offers a transparent, extensible tool for computational social science in an era of shifting digital ecosystems.
\end{abstract}

\subsubsection*{Keywords:}

Narrative analysis, Sentiment analysis, Topic modeling, Social media mining, Decentralized social media

\section{Introduction}

Social media platforms play an increasingly central role in shaping and reflecting public discourse. Historically, Twitter (now X) has served as a core dataset for researchers studying online behavior, public opinion, and effective communication \parencite{bruns2013twitter, kwak2010twitter, Das2019_ExtractingPatternsFromTwitter, DasDutta2020_CharacterizingPublicEmotions, Das2021a}. However, recent API restrictions \parencite{cip2023twitterapi} have severely limited access for academic and open-source communities. As a result, attention has shifted toward decentralized platforms such as Bluesky \parencite{bluesky2023overview}, which offer greater data openness and user autonomy.

While historical datasets like the Twitter Stream Grab remain valuable, many tools built around them are outdated or constrained in functionality. Tools like TweetViz, TExVis, and Gephi provide partial solutions but lack the scalability, usability, or analytical depth necessary for contemporary narrative analysis \parencite{al2024twixplorer}. TwiXplorer \parencite{al2024twixplorer} advanced the field by combining sentiment analysis, topic modeling, and visualization in an interactive dashboard, but remains limited to static archives of Twitter (X) data.

\textbf{CognitiveSky} extends this paradigm into the real-time and decentralized space. As shown in Table~\ref{tab:cs-vs-twixplorer}, it builds upon the principles demonstrated in TwiXplorer, introducing a modular, open-source framework for ingesting, labeling, and summarizing Bluesky discourse. Unlike static analysis tools, CognitiveSky operates continuously on live data, supports transformer-based sentiment and emotion detection, and produces structured JSON summaries for interactive visualization.

\begin{table*}[thb]
\centering
\caption{\label{tab:cs-vs-twixplorer} Comparison of CognitiveSky and TwiXplorer in terms of architectural and functional capabilities.}
\vskip 3pt
\begin{tabular}{
  >{\raggedright\arraybackslash}p{3cm} 
  >{\raggedright\arraybackslash}p{6cm} 
  >{\raggedright\arraybackslash}p{6cm}
}
\hline
\textbf{Feature} & \textbf{CognitiveSky} & \textbf{TwiXplorer \parencite{al2024twixplorer}} \\
\hline
Platform Support & Bluesky (real-time, decentralized) & Twitter (static JSON archives) \\
Ingestion Method & WebSocket stream from Firehose & Manual upload of historical datasets \\
Labeling Models & Transformer-based (sentiment, emotion, topic) & term frequency–inverse document frequency (TF-IDF) with semantic projection \\
Topic Modeling & MiniBatchNMF on TF-IDF vectors & TF-IDF clustering with centroids \\
Visualization & Modular dashboard (Next.js + Recharts) & Semantic maps, timeline plots \\
Deployment & Fully open-source, cloud-ready (free-tier) & Web interface with limited backend details \\
Use Case Focus & Real-time monitoring of mental health discourse & Analysis of historical social/political datasets \\
Reproducibility & continuous integration (CI) pipelines, snapshot hashing, JSON exports & No published automation or reproducible pipeline \\
\hline
\end{tabular}
\end{table*}

CognitiveSky is designed to be lightweight, low-cost, and scalable. Its modular architecture supports adaptation across a range of domains; from mental health discourse, where emotionally expressive and time-sensitive narratives are common \parencite{guntuku2017detecting,de2013predicting}, to areas like disinformation detection, civic engagement, and crisis response. In the sections that follow, we describe CognitiveSky’s system architecture, Natural Language Processing (NLP) pipeline, summarization engine, and interactive dashboard, along broader applicability.

\section{System Overview}

\begin{figure*}
    \centering
    \includegraphics[width=1\linewidth]{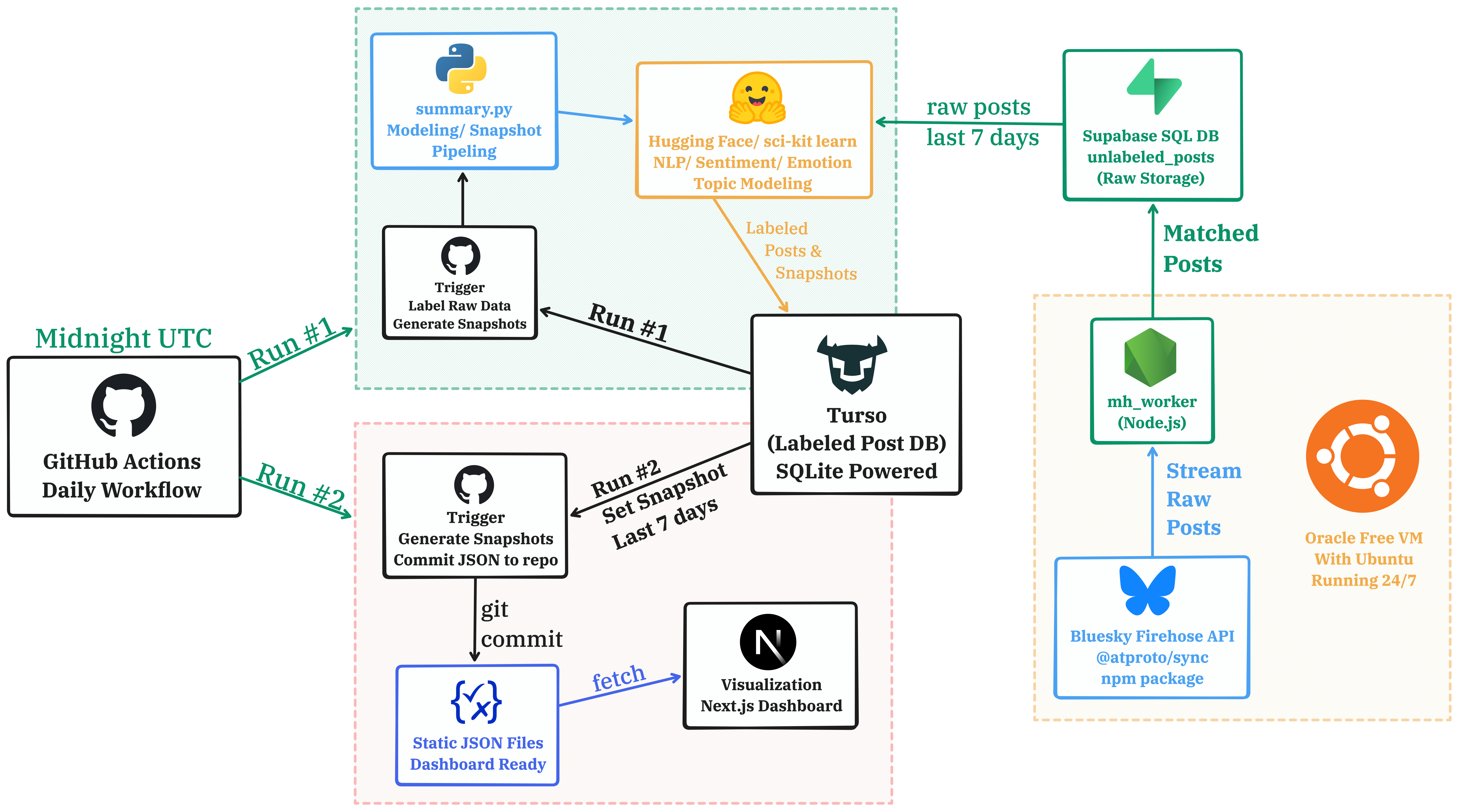}
    \caption{End-to-end overview of CognitiveSky’s modular pipeline.}
    \label{fig:architecture}
\end{figure*}

CognitiveSky is structured as a fully open-source modular pipeline\footnote{The full source code including ingestion, labeling pipeline, interactive dashboard, and, GitHub Actions based automation is available at \url{https://github.com/gauravfs-14/CognitiveSky}, licensed under MIT.}for streaming, labeling, enriching, and visualizing discourse data from the decentralized Bluesky platform. It is designed for real-time operation with support for daily summarization and long-term trend analysis. The system is divided into three loosely coupled layers: ingestion, labeling, and summarization (see Figure~\ref{fig:architecture}).

\subsection{Ingestion and Preprocessing}

The ingestion pipeline operates 24/7 on a lightweight Oracle Cloud Virtual Machine \parencite{oraclecloudfree} using Node.js. It connects to the AT Protocol Firehose via WebSocket \parencite{kleppmann2024bluesky} and listens for \texttt{app.bsky.feed.post} events, which represent newly published public posts on the Bluesky network. To filter for relevant content, CognitiveSky applies a keyword-based heuristic using array of English-language expressions as keywords. These patterns are compiled into case-insensitive regular expressions and evaluated in real time using a JavaScript-based matcher.

Posts that pass the filter are transformed into structured objects containing metadata such as post ID, user decentralized identifiers (DID), timestamp, raw text, and optional fields like language tags and embedded media. These records are held in an in-memory buffer and batch-inserted into Supabase \parencite{supabase2025} every 5 seconds or after 200 posts, with a retry mechanism that discards records after three failed attempts.

Supabase acts as a transient staging layer between ingestion and labeling. A graceful shutdown handler ensures buffered data is flushed to persistent storage on termination. Unlike traditional pipelines, CognitiveSky defers cleaning to the annotation stage, where text is normalized, URLs and mentions removed, and posts validated for minimum content before NLP processing.

\subsection{Data Labeling}

Posts in the Supabase buffer are processed by a labeling pipeline written in Python, triggered daily by a GitHub Action  workflow \parencite{githubactions}. This component loads posts in batches of up to 64, enabling safe memory usage in CI environments. It assigns each post affective and semantic labels using transformer models and topic clustering.

Sentiment classification is performed using a RoBERTa-based model from CardiffNLP \parencite{cardiffnlp2021}, assigning one of three polarities: \textit{positive}, \textit{neutral}, or \textit{negative}. Emotion detection follows using a DistilRoBERTa model trained on the GoEmotions dataset \parencite{hartmann2022}, which identifies categories like \textit{joy}, \textit{fear}, and \textit{anger}. For each post $x$, a contextual embedding is extracted from the special classification token, denoted as \texttt{[CLS]}, which represents an aggregated summary of the input sequence. This embedding is passed through a softmax classifier:

\[
\hat{y} = \mathrm{softmax}(W \cdot \mathrm{CLS}(x) + b)
\]

These models are optimized for short-form, informal text and trained using cross-entropy loss. Intermediate memory is explicitly released between batches to avoid exhaustion in constrained CI contexts.

Following the emotion detection, the textual content of each post is vectorized using the TF-IDF method \parencite{salton1988term} to capture salient lexical features. Topic clustering is subsequently performed using MiniBatch Non-negative Matrix Factorization \parencite{lee1999learning}, which decomposes the TF-IDF matrix to assign each post to a dominant topic (e.g., \textit{topic 2}). For interpretability, the most representative keywords for each topic cluster are extracted and retained for visualization in the dashboard.

Final annotations are stored in a Turso (libSQL) database \parencite{tursodb2025} using an insert or ignore policy to prevent duplication. Once inserted, the corresponding records are purged from Supabase, preserving pipeline throughput and preventing redundant processing.

\subsection{Real-time Interactive Dashboard}

\begin{figure*}[htb]
  \centering
  \begin{minipage}[t]{0.23\textwidth}
    \centering
    \includegraphics[width=\linewidth]{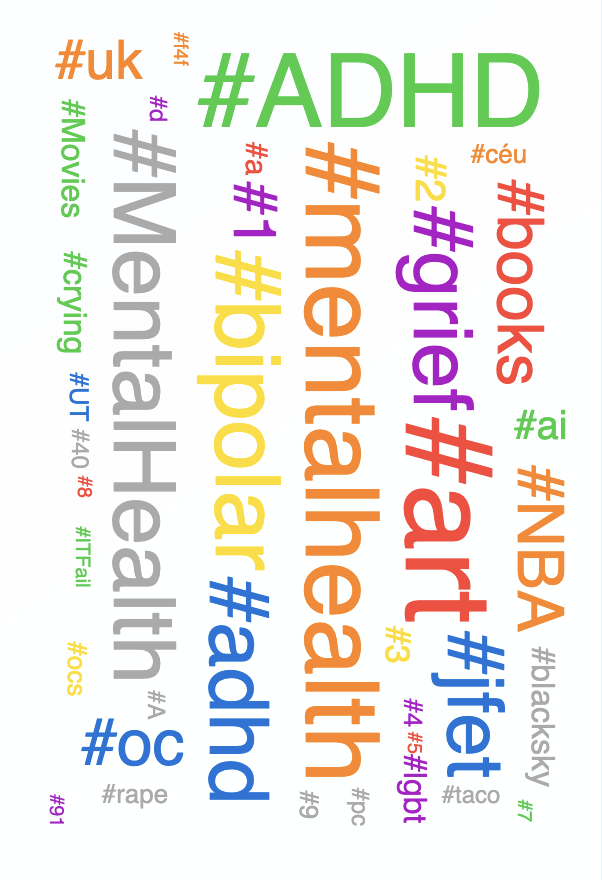}
    \subcap{(a) Hashtag Cloud}
    \label{fig:dash-hashcloud}
  \end{minipage}
  \hfill
  \begin{minipage}[t]{0.23\textwidth}
    \centering
    \includegraphics[width=\linewidth]{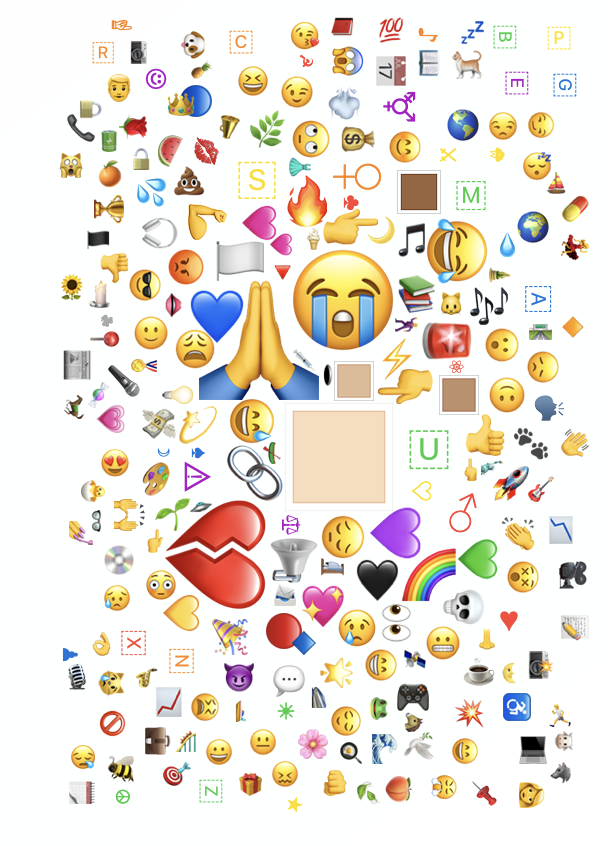}
    \subcap{(b) Emoji Cloud}
    \label{fig:dash-emojicloud}
  \end{minipage}
  \hfill
  \begin{minipage}[t]{0.23\textwidth}
    \centering
    \includegraphics[width=\linewidth]{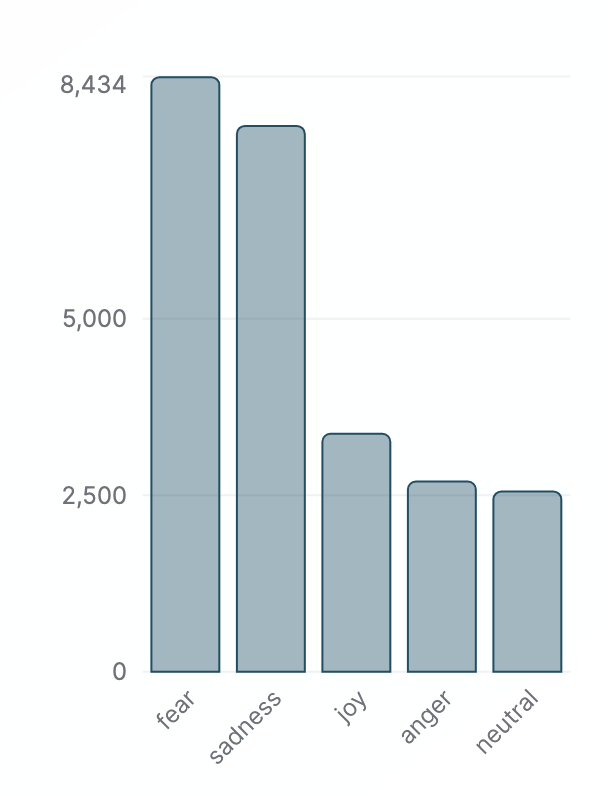}
    \subcap{(c) Emotion Graph}
    \label{fig:dash-emotiongraph}
  \end{minipage}
  \hfill
  \begin{minipage}[t]{0.23\textwidth}
    \centering
    \includegraphics[width=\linewidth]{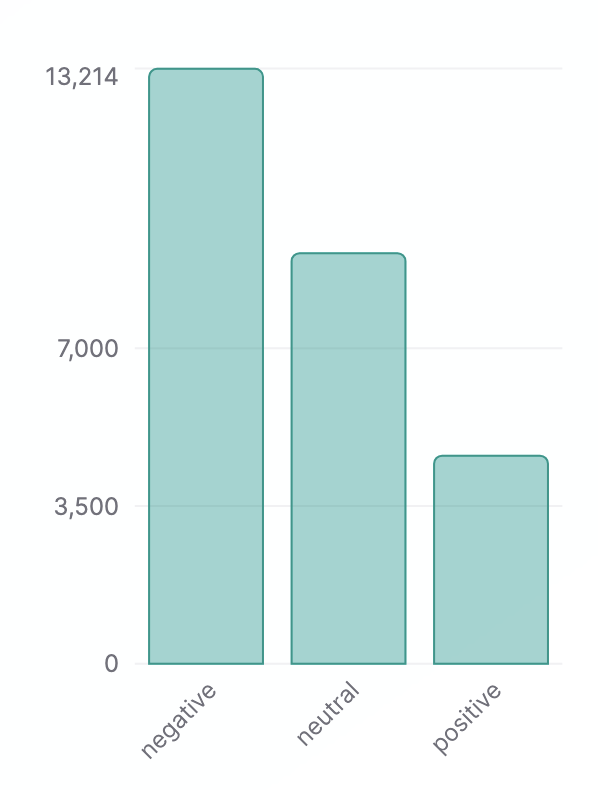}
    \subcap{(d) Sentiment Graph}
    \label{fig:dash-sentimentgraph}
  \end{minipage}

  \vskip 1em

  \begin{minipage}[t]{1\textwidth}
    \centering
    \includegraphics[width=\linewidth]{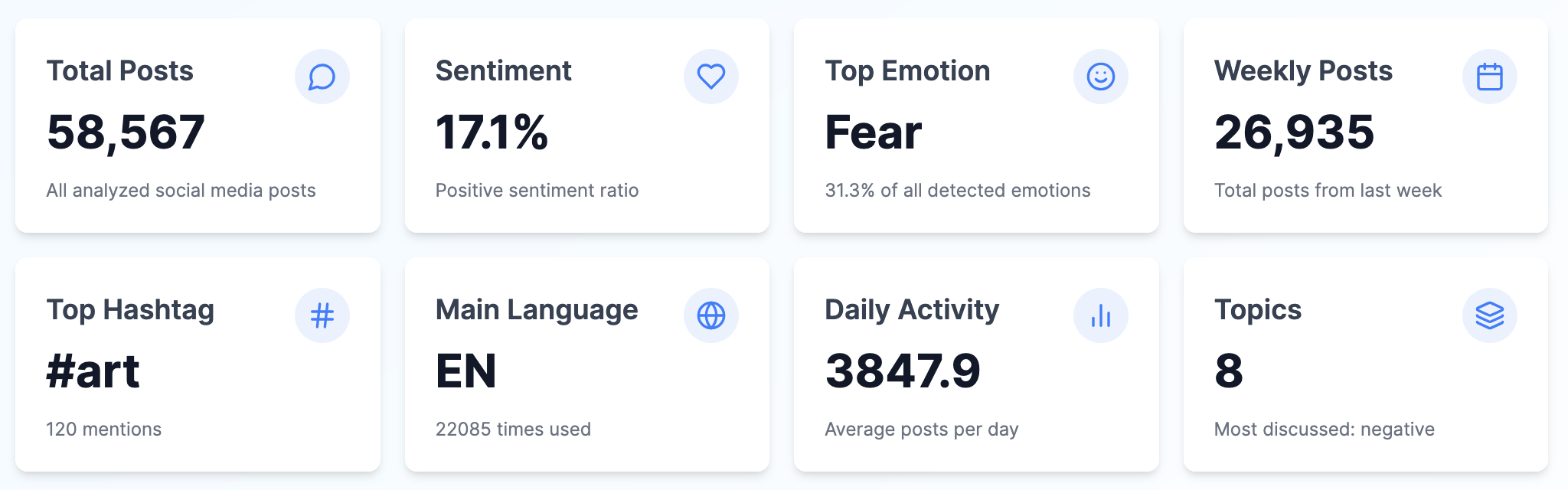}
    \subcap{(e) Dashboard Overview Panel}
    \label{fig:dashboard-quickstat}
  \end{minipage}

  \caption{Interactive visual components from the CognitiveSky dashboard. Panels include (a–b) word clouds for hashtags and emojis, (c–d) bar charts for emotions and sentiment, and (e) a snapshot of key summary metrics.}
  \label{fig:dashboard-ui}
\end{figure*}

To support time-aware and interpretable analysis, CognitiveSky aggregates labeled records from the Turso database into structured JSON summaries\footnote{Default configuration supports the past 7 days; longer windows can be configured as needed.}. These snapshots, generated daily and weekly, encode the temporal, affective, and topical dimensions of the dataset.  Snapshot generation combines SQL-based querying with in-memory aggregation routines. Temporal summaries capture daily counts of posts, sentiment shifts, and emotional variation. Topic summaries include cluster frequency, representative keywords, and associated emotional context. Hashtags and emojis are extracted using regular expressions, with co-occurrence graphs built from within-post pairings. A dedicated meta file tracks global metrics such as total posts, unique users, and top-used terms.

CognitiveSky’s dashboard functions as the primary interface for exploring mental health discourse on Bluesky. Built with Next.js and Recharts and deployed via Vercel\footnote{\url{https://cognitivesky.gaurabchhetri.com.np/}}, it is powered entirely by static JSON snapshots, allowing low-latency access, serverless architecture, and zero dependency on live database queries.

The dashboard is modular by design, with views mapped directly to individual snapshot files. The landing page presents a holistic overview of dataset activity, including post volume, top emotions, trending hashtags, and language usage (Figure~\ref{fig:dashboard-ui} illustrates some components of the interface). Users can navigate to specialized modules for sentiment and emotion trends, topic clusters, hashtag dynamics, emoji usage, and a timeline-based comparison interface.

Interactivity features such as animated charts, tool-tips, and responsive layout ensure usability across devices. This lightweight yet information-rich design makes the dashboard a powerful tool for both exploratory analysis and longitudinal monitoring of online mental health narratives.

CognitiveSky is developed with an emphasis on ethical responsibility and privacy preservation in the processing of social media data. The system is explicitly designed to ingest and analyze only content that is publicly accessible and ethically permissible for research purposes. To that end, CognitiveSky exclusively utilizes non-reply, public-facing posts from the Bluesky Firehose. This approach aligns with prevailing ethical standards concerning the use of publicly available data for computational analysis. CognitiveSky references user accounts solely via DIDs. These DIDs are intrinsic to the Bluesky protocol and are not enriched with or mapped to any external identity sources or platforms.  Importantly, the system does not display or expose individual-level posts at any point in the user interface, analytic dashboard, or in downloadable datasets. All textual data is retained and processed exclusively for the purposes of aggregate statistical and thematic analysis. Posts are abstracted into categories such as topic clusters, sentiment distributions, or engagement patterns, with no individual post text, metadata, or identifiers ever being surfaced or shared. 

\section{Technology Optimizations}

\begin{table*}[thb]
\centering
\caption{\label{tab:techstack} End-to-end technology stack used in CognitiveSky, with functional roles and rationale.}
\vskip 3pt
\begin{tabular}{
  >{\raggedright\arraybackslash}p{3cm} 
  >{\raggedright\arraybackslash}p{4cm} 
  >{\raggedright\arraybackslash}p{8cm}
}
\hline
\textbf{Layer} & \textbf{Technology} & \textbf{Purpose and Rationale} \\
\hline
Ingestion & Node.js + PM2 on Oracle & Handles Firehose streaming using asynchronous logic; low-cost continuous operation. \\
Storage (Raw) & Supabase (PostgreSQL) & Temporary buffer for ingested posts; REST API access and integration ease. \\
Labeling Pipeline & Python + GitHub Actions & Automates daily transformer-based labeling; ensures reproducibility through CI workflows. \\
NLP Models & HuggingFace Transformers & State-of-the-art models fine-tuned on social data; robust sentiment and emotion inference. \\
Storage (Labeled) & Turso (libSQL) & Serverless, fast-access SQL storage; optimized for deduplication and querying. \\
Visualization & Next.js + Recharts on Vercel & Frontend for dashboard; serverless, statically deployed using precomputed JSONs. \\
\hline
\end{tabular}
\end{table*}

CognitiveSky is architected as a lightweight, modular, and reproducible pipeline optimized for real-time decentralized data analysis. Each layer of the system, from ingestion to labeling to visualization, was designed to operate autonomously using only open-source and free-tier infrastructure. At core, we are utilizing JavaScript, TypeScript for data collection and visualization, a Python-based pipeline for data labeling and processing, and GitHub actions as a CI layer, enabling automated daily running of the Python-based pipeline. An overview of the technologies used across each layer of the system is presented in Table~\ref{tab:techstack}. The table summarizes the selected tools, platforms, and frameworks along with their function and justification in the overall pipeline.

This stack enables CognitiveSky to function continuously and autonomously, making it suitable for research groups operating with limited infrastructure. It also ensures transparency, extensibility, and reproducibility, all of which are critical for community-driven development and open social data science.

\begin{table*}[thb]
\centering
\caption{\label{tab:where_what_how} System-level challenges encountered in CognitiveSky and corresponding engineering resolutions.}
\vskip 3pt
\begin{tabular}{
  >{\raggedright\arraybackslash}p{2.5cm} 
  >{\raggedright\arraybackslash}p{5.5cm} 
  >{\raggedright\arraybackslash}p{7cm}
}
\hline
\textbf{Where} & \textbf{Challenge} & \textbf{Solution} \\
\hline
Ingestion (Oracle VM) & Maintaining 24/7 streaming on constrained hardware & Lightweight async Node.js worker deployed with PM2 for autorestart and memory safety. \\
Supabase (Raw Buffer) & Accumulation of unlabeled posts impacting performance & Used Supabase as a temporary buffer; migrated data daily to Turso and purged staged entries. \\
GitHub Actions & Timeouts and memory limits in CI jobs & Batched processing, cached models, and memory cleanup between steps to ensure stability. \\
Transformer Inference & High memory usage for labeling models & Switched to smaller transformers with CPU-safe inference and tuned batch sizes. \\
Turso Migration & Duplicate inserts or lost entries & Used \texttt{INSERT OR IGNORE} with transaction logs and batch-level error tracking. \\
Snapshot Generation & Recomputing unchanged daily summaries & Employed SHA-256 hashing to skip snapshots with identical content. \\
Snapshot Growth & Dashboard slowdown from bloated summary files & Limited exports to a 7-day rolling window and modularized summary outputs. \\
\hline
\end{tabular}
\end{table*}

\section{Use Case: Mental Health Narratives}

Although CognitiveSky is designed as a general-purpose framework for analyzing decentralized social media discourse, its initial deployment focuses on the domain of mental health. The decision to prioritize mental health stems from the growing concern about the rise of anxiety, depression, and emotional distress expressed online, particularly among younger populations who are early adopters of platforms like Bluesky \parencite{pew2025teens}.

Mental health–related discussions on social media, like other sensitive topics, often carry heightened emotional intensity and subtle contextual cues, with implicit signals that may point to underlying distress or calls for help requiring timely recognition and response \parencite{chancellor2020methods, Das2024c}. These characteristics make them an ideal testbed for CognitiveSky's NLP-driven labeling and visual analysis pipeline. The system enables researchers to capture not just what users are discussing, but how they feel, when they engage, and what kinds of language or cultural markers they use to express emotional states.

In this use case, CognitiveSky continuously monitors the Bluesky Firehose for public posts containing mental health-related keywords such as “depression,” “therapy,” “panic attack,” “burnout,” or “healing.” These posts are streamed in real time to Supabase and then labeled using transformer models to assign sentiment and emotion categories. Once labeled, the posts are analyzed for daily trends, emotional composition, and associated hashtags or emojis.

\section{Replicable Framework and Other Use Cases}

While initially implemented for mental health narrative, CognitiveSky's modular, scalable design allows it to adapt seamlessly to a variety of real-world scenarios beyond mental health. Its ability to extract structured insights from unstructured, real-time, decentralized social media data enables use in critical areas like information security, crisis detection, social movement tracking, and academic inquiry.

The growing prevalence of disinformation, civic polarization, and crisis events on social media calls for tools that go beyond surface-level metrics. CognitiveSky enables layered narrative detection by combining clustering, sentiment, and emotion analysis on decentralized platforms. In disinformation and coordinated influence campaigns \parencite{ng2023coordinated}, CognitiveSky helps identify clusters of semantically similar posts with affective manipulation, such as spikes in fear, anger, or outrage. These patterns often indicate conspiracy narratives, hate speech, or influence operations, aiding digital literacy efforts and ethical governance. During emergencies, like natural disasters, pandemics, or social unrest, CognitiveSky supports real-time monitoring of emotionally charged language and evolving topics \parencite{lokmic2023lessons}. Its lightweight infrastructure makes it suitable for rapid deployment by NGOs, grassroots groups, or government agencies, even without commercial APIs or large-scale compute resources. In civic contexts, CognitiveSky enables fine-grained analysis of public sentiment and emotional tone during movements (e.g., \#MeToo, \#BLM), elections, and legislative debates \parencite{zhang2025individual}. With optional metadata (e.g., geolocation), it supports regional and demographic comparisons for researchers, journalists, and civic technologists.

CognitiveSky is designed for transparent, reproducible, and interdisciplinary research. It can support longitudinal discourse analysis for social scientists, semantic evolution studies for linguists, and policy-response tracking for public health researchers. The pipeline’s modular design and automation through GitHub Actions and open-source models lower the barrier for teams with limited infrastructure. Its versatility allows researchers to explore diverse questions without high overhead.

\section{Conclusions}

We introduced CognitiveSky, an open-source, real-time system for detecting, labeling, and visualizing discourse trends on the decentralized Bluesky network. Built with modularity, reproducibility, and accessibility at its core, CognitiveSky bridges the gap between large-scale data ingestion and structured narrative analysis. By applying state-of-the-art NLP models to live social streams and pairing results with lightweight summarization and visualization, the system enables timely interpretation of public sentiment, emotion, and topic dynamics. A mental health case study highlighted its ability to capture nuanced emotional expressions and evolving themes within online communities.

CognitiveSky’s architectural choices (e.g., free-tier deployment, serverless infrastructure, static dashboarding, and modular pipelines) allow it to operate without costly infrastructure. This makes the system particularly well-suited for researchers, civic organizations, and journalists who seek to investigate discourse on decentralized platforms. However, limitations remain: the current pipeline relies on English-only models, real-time scalability is constrained by rate limits and API policies, and static dashboards restrict interactivity and customization. Moreover, data volatility on decentralized networks presents challenges for replicability and long-term validation.

Future work should focus on multilingual and cross-cultural expansion, advanced narrative-shift detection, semantic clustering for more robust topic tracking, and user-centric longitudinal analysis. Interactive and adaptive dashboards could enhance transparency while supporting richer exploration of results. With these developments, CognitiveSky can evolve into a scalable, ethical, and reproducible tool for understanding discourse dynamics across an increasingly diverse range of decentralized ecosystems.

\printbibliography

\end{document}